# MS-Net: A Multi-modal Self-supervised Network for Fine-Grained Classification of Aircraft in SAR Images

Bingying Yue, Jianhao Li, Hao Shi*, Yupei Wang, Honghu Zhong
Beijing Institute of Technology

*Abstract*—Synthetic aperture radar (SAR) imaging technology is commonly used to provide 24-hour all-weather earth observation. However, it still has some drawbacks in SAR target classification, especially in fine-grained classification of aircraft: aircrafts in SAR images have large intra-class diversity and inter-class similarity; the number of effective samples is insufficient and it's hard to annotate. To address these issues, this article proposes a novel multi-modal self-supervised network (MS-Net) for fine-grained classification of aircraft. Firstly, in order to entirely exploit the potential of multi-modal information, a two-sided path feature extraction network (TSFE-N) is constructed to enhance the image feature of the target and obtain the domain knowledge feature of text mode. Secondly, a contrastive self-supervised learning (CSSL) framework is employed to effectively learn useful label-independent feature from unbalanced data, a similarity perception loss (SPloss) is proposed to avoid network overfitting. Finally, TSFE-N is used as the encoder of CSSL to obtain the classification results. Through a large number of experiments, our MS-Net can effectively reduce the difficulty of classifying similar types of aircrafts. In the case of no label, the proposed algorithm achieves an accuracy of 88.46% for 17 types of aircraft classification task, which has pioneering significance in the field of fine-grained classification of aircraft in SAR images.

## I. Introduction

Synthetic Aperture Radar (SAR) is not restricted by illumination and complex weather conditions, and has the ability of surface penetration. With the help of aircraft, satellite and other flight platforms, SAR can perform 24-hour all-weather observation over a large area. With these advantages, SAR has unique preponderance in various fields such as disaster monitoring, environmental monitoring, ocean monitoring, resource exploration, and so on [1], [2]. However, due to geometric transformations such as speckle and overlay in SAR images, the imaging results are not intuitive. Therefore, in recent years, SAR image interpretation has become a hot topic among scholars from all over the world [3].

Target recognition is a pivotal study of SAR image interpretation. The fine category of the target can be identified via feature extraction and analysis of the image. According to whether the features need to be constructed manually, there are two main methods in current research: traditional methods and deep learning methods.

Traditional SAR target classification methods can be mainly divided into two categories: template-based [4]-[8] and model-based [9]-[11]. The template-based method mainly adopts the approach of "feature extraction and classifier". Its key lies in feature extraction and selection, which requires profound domain knowledge as the basis. Typical features include geometric feature, texture feature, electromagnetic scattering feature and transform domain feature. This method is intuitive and easy to understand. However, due to the complex imaging conditions of SAR images and their susceptibility to clutter, constructing an effective feature database based on engineering experience is necessary for this method. It's difficult to ensure the feature robustness. The core of the model-based approach lies in the design of the target's model. By the physical modeling of the target, the scattering process of electromagnetic wave irradiation on the target is simulated, and the features of the target are predicted under different attitude, configuration and environmental conditions. However, the parameter information of the target is required in advance. Besides, with the enhancement of the resolution, the calculation amount of physical modeling and the difficulty of simulation increase accordingly, which make it impossible to meet the demands of efficiency in practical applications.

In recent years, deep learning has been employed to SAR target interpretation due to its robust automatic feature extraction capability [12], [13]. Compared with traditional methods, deep learning technology has the advantage of automatically learning effective features from data, and is expected to extract features that can significantly surpass artificial design in classification performance. Fine-grained classification of aircraft in high-resolution SAR images determines the specific type of aircraft by analyzing the salient features in SAR slices. Compared with traditional SAR image classification, the fine-grained classification is more challenging owing to the large intra-class diversity and inter-class similarity of SAR targets. One target can appear significantly different under different conditions, such as resolution, azimuth angle, polarization mode and other parameters and background. Nonetheless, the size and shape of targets within different subcategories may be similar. To manage these confusing images, the network needs to recognize subtle differences between different types of objects. Similar targets are almost indistinguishable if you only depend upon images. In the fine-grained classification task, apart from a few popular fine-grained approaches that focus on distinguishing regions of the image [14], [15], multi-branch learning [16], [17] or specific data enhancement [18], [19] another feasible idea is to introduce the additional information. Lu et al. [20] proposed a method that incorporates text modal descriptions with visual modal features. The integration of these features results in a significant decrease in misclassification rates for new categories. Zhao et al. [21] proposed a feature balancing strategy to effectively utilize image super resolution features. Image super resolution features can provide auxiliary structural information to improve visual tasks. Song et al. [22] use auxiliary language patterns to improve fine-grained classification, exploring knowledge outside the visual domain of language modes in the interactive alignment paradigm. In addition to images, multi-modal information can effectively improve the accuracy of the classification network.

Moreover, the large annotated datasets are essential for the development of deep learning. ImageNet dataset [23], one of the most classic datasets in the field of optical image classification, contains 14,197,122 images, distributed over 21,841 categories. Compared with optical images, obtaining high-resolution SAR images is costly. Currently, the relevant studies that can be retrieved are carried out mostly rely on self-built datasets. Therefore, insufficient data volume is the major limitation for using deep learning algorithms in SAR automatic target recognition (ATR) tasks. Due to the complex structure and the large number of parameters of deep neural network, the direct training of neural network under the condition of limited training samples is likely to cause serious overfitting. Current research directions of deep learning technology for image classification include data extension, model optimization, small sample learning and self-supervised learning (SSL) [24].

At present, the number of high-resolution SAR images available has turned a corner, but it is difficult to add accurate and complete annotations to the acquired data [25]. In recent years, various countries have launched SAR satellites with excellent technology, such as medium-orbit, high-resolution and widefield Marine SAR satellites, which can acquire SAR images covering an area of nearly one million square kilometers every day. In addition, with the resolution of SAR sensor reaching sub-meter level, aircraft target recognition based on high resolution remote sensing images has gradually developed from simple target detection and positioning to target specific type recognition, and the need for intelligent SAR image interpretation is extremely urgent. Although the amount of available data has been improved, SAR images are obtained by compressing the range and azimuth-direction of backscattered echoes, and often have geometric transformations such as speckle and overlay, which make the imaging results less intuitive. Thus, SAR interpretation depends on expert knowledge. The cost of labeling is very high. Currently, there are few publicly available aircraft model classification datasets.

SSL is an important branch of unsupervised learning. The model directly learns a feature extractor from unlabeled data, which is suitable for scenarios where the acquisition of annotated information is costly. Therefore, SSL can flexibly acquire the diversified features contained in remote sensing images, thus greatly reducing the dependence on annotated data.

Based on above discussions, the previous work on fine-grained classification of aircrafts in SAR images still has following unsolved problems: Due to SAR's imaging mechanism based on electromagnetic scattering, aircrafts in SAR images have relatively high intra-class diversity, and aircrafts of the same category have high inter-class similarity; the number of effective samples is limited, and the labeling requires much expert knowledge. These factors are easy to interfere with the classification model. This article will further explore and improve the effectiveness of deep neural network. Main contributions are summarized below:

1) Two-sided path feature extraction network (TSFE-N) is built to fully explore the possibilities of multi-modal information. The network contains two key components: size information extraction branch (SIEB) and self-attention enhancement module (SAEM). SIEB is introduced to extract size information to obtain the domain knowledge features of text mode. SAEM is designed to enhance image features, the global information features of targets are refined by integrating convolution and self-attention paradigms.
2) In order to effectively learn useful label-independent initializing features from unbalanced data, a CSSL framework is adopted. We design a similarity perception loss function (SPloss) to balance the different categories and improve the generalization ability of the network.

The rest of this article is organized as follows. Section II introduces the related work. Section III describe the proposed fine-grained aircraft classification algorithm. Section IV elaborates on the implementation of the experiment and the performance of the proposed method. Section V discusses the success of our method and future research prospects in a broader context. Section VI concludes the study.

## II. RELATED WORK

### A. CNN-Based SAR Target Interpretation

Since AlexNet [26] was proposed in 2012, Convolutional Neural Networks (CNNs) have become a mainstream method in the field of image classification in natural scenes. By virtue of their powerful feature extraction ability, CNNs have gradually gained popularity in the SAR image processing field. Since deep neural networks showed excellent performance in ship and vehicle target interpretation in SAR images, they were progressively applied in SAR aircraft detection and recognition. Because of the distinctive SAR imaging system and the complicated geometric structure of the target, it is necessary to improve the algorithm in combination with the imaging characteristics of the aircraft when using deep neural network to interpret SAR images. Diao et al. [27] presented a target prepositioning algorithm based on Constant False Alarm Rate (CFAR). This method combined the intensity of position-regression based CNN framework with the significant features of the target, making the match of the target scale more accurate. Guo et al. [28] proposed a hybrid method based on scattering information enhancement and attention pyramid network to deal with aircraft's dispersion and variability. This method achieved good detection effect on GF-3 aircraft dataset with 1m resolution. Zhao et al. [29] proposed a feature refinement and alignment network based on attention mechanism. The discrete features of aircraft were collected by dilated convolution branches with different expansion rates, and feature refinement and alignment were realized by attention mechanism and deformable convolution. The effectiveness of the algorithm was verified on GF-3 and 0.5m TerraSAR-X satellite images. Kang et al. [30] designed a scattering feature relation network for imagery to ensure the completeness of detection results by enhancing the relationships between scattering points. A remarkable fusion module was designed to adaptively combine the features of several layers and enhance the recognition of the features. Zhao et al. [31] proposed a global instance level comparison module to improve inter-class differences and intra-class compactness, which demonstrated good classification and localization accuracy in the Gaofen-3 SAR aircraft detection dataset. Kang et al. [32] proposed a scattering topology network in SAR images. This network exploits imaging variability and component scattering discreteness from different angles. The network models the spatial relationship and semantic information interaction of scattering points, and achieves accurate classification on the SAR-ACD dataset. In the above approach, improving networks in the traditional deep learning models can often be challenging when faced with tasks of inconsistent categories. When patterns vary significantly from one category to another, it is difficult for models to learn consistent representations. This will lead to overfitting or underfitting, which ultimately limit the models' generalization ability across different datasets.

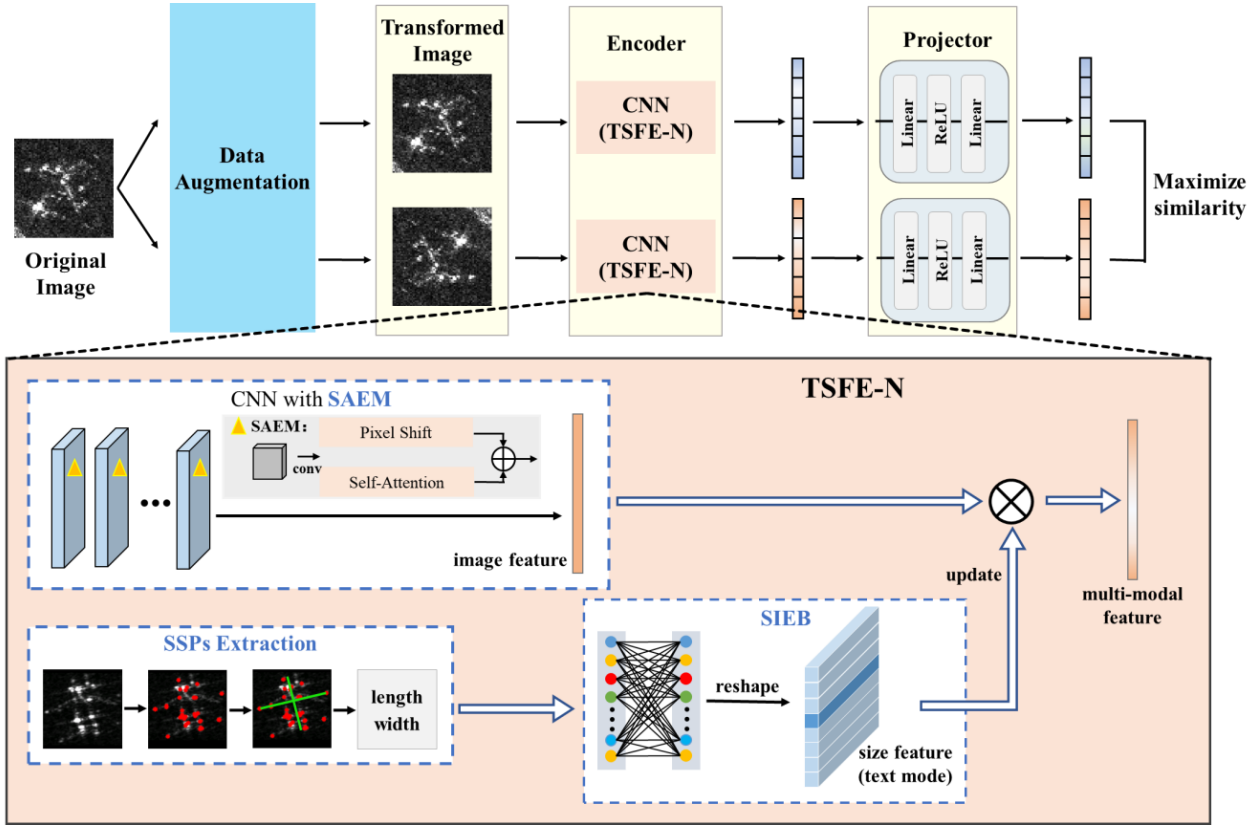

**Fig. 1.** Overall structure of the proposed MS-Net

Faced with the issue of inadequate SAR data, the direct network migration from natural scenes is often unable to achieve satisfactory performance. Huang et al. [25] thoroughly explored the application of migration learning methods to SAR images, including which network and source tasks are more suitable for migration to SAR, selection of intermediate layer for effective feature transmission, and efficient migration methods for target recognition in SAR images. They also proposed a domain adaptive transfer and migration method based on multi-source data to minimize the gap between source data and SAR target. Zhang et al. [33] proposed a domain knowledge driven dual-flow depth network that integrated SAR domain knowledge related to azimuth angle, amplitude, and phase data. Sun et al. [34] combined the scattering features in the SAR imaging process with the neural network, and designed a meta-learning network for imaging variability, which integrated angle adaptive classifier and the frequency-domain information embedded module. However, the effectiveness of transfer learning highly depends on the similarity between the source domain and the target domain. If the two domains have significant differences, the method's efficacy may be limited. Small sample learning may overly focus on a limited amount of data and disregard the overall distribution of data samples, ultimately leading to the overfitting problem in the model.

Different types of aircrafts come in different sizes. The length of ordinary fighter jets is about 15-20m, but the wingspan of the largest transport aircraft can reach 88m. This means that, at the same resolution, the pixel range of different aircrafts types in remote sensing images is quite different. Similarly, different types of aircrafts have different aspect ratios. Therefore, to leverage the latent impact of extra information, the size information representation of text mode helps obtain high-dimensional interactions between features.

*B. SSL-Based Object Classification*

SSL methods can be categorized into two main types: generative self-supervised network (GSSL) and contrastive self-supervised network (CSSL) [35]. GSSL learns key image features by repairing and restoring artificially damaged images, while CSSL learns image features by comparing the loss function to distinguish positive and negative samples, constantly narrowing the distance among positive samples and enlarging the distance among negative samples [36]. Accordingly, GSSL focuses heavily on pixel information, which may cause resource redundancy. CSSL is widely used by researchers [37] as it focuses on extracting features that are distinguishable among categories, and it saves more memory than GSSL. According to the presence or absence of negative samples, CSSL methods can be divided into two types. Typical negative sample CSSL methods include MoCo [38], SimCLR [39], CMC [40], while typical non-negative sample CSSL methods include BYOL [41] and SimSiam [42]. CSSL pre-training neural network can be used for downstream supervision tasks [43]. Therefore, CSSL can be used to build a pre-training model on untagged data. The key idea of CSSL is to keep narrowing the range among positive samples, as well as widening the range among negative samples. Table I shows the comparison of prevailing CSSL cla-ssifiers. The basic goal of momentum comparison is to maintain the dictionary as a collection of data samples in a queue. Each self-supervised network model has its benefits and drawbacks. For the SAR aircraft classification task studied in this article, the choice of the mainstream model should be experimentally evaluated based on the existing dataset size and computing resources.

TABLE I
COMPARISON OF PREVAILING CSSL CLASSIFIERS.

| Method | Negative pairs | Momentum encoder |
|---|---|---|
| SimCLR | ✓ | - |
| MoCo | ✓ | ✓ |
| MoCo v2 | ✓ | ✓ |
| BYOL | - | ✓ |
| SimSiam | - | - |

III. METHODOLOGY

The overall structure of the proposed MS-Net is illustrated in Fig. 1. Firstly, multi-modal knowledge features are extracted by TSFE-N. TSFE-N contains two primary components, SIEB and SAEM. SIEB updates the targets' scale features of text mode to image features through adaptive projection, so as to realize information interaction in a higher and wider dimension. SAEM performs self-attention enhancement operations on the image features, this focuses the image representation on the optimal attention. Secondly, a CSSL network based on similarity perception is employed to reduce the model's dependence on labeled samples and help to learn useful initializing information from the unevenly distributed dataset. Finally, TSFE-N is adopted as the encoder for feature extraction in the CSSL.

*A. Two-sided Path Feature Extraction Network (TSFE-N)*

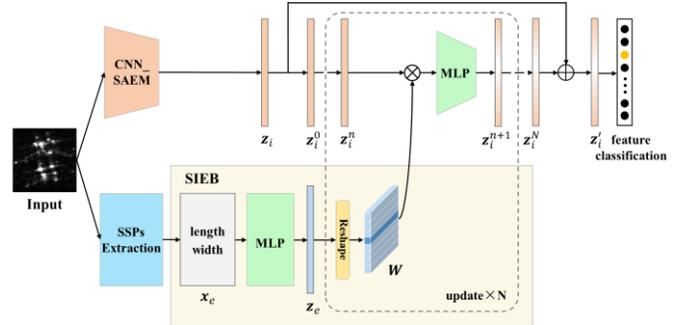

**Fig. 2.** Structure of the proposed TSFE-N.

Fig. 2 depicts the structure of the proposed TSFE-N, which is used to extract multi-modal features from images. SIEB and SAEM are two key components of TSFE-N. First, the aircraft's size information of text mode is obtained by extracting strong scattering points (SSPs) from target slices. SIEB is designed to obtain the feature attributes of domain knowledge from SSPs. To be specific, SIEB updates the target size information features of text mode to image features through adaptive projection, so as to realize efficient and high-level information interaction. Similar image features are mapped into different positions in feature space via dynamic and instantiated projection of target dimensional features. SAEM is created to enhance the features and refine the global features of the targets. Different from the traditional convolution kernel limits the size of the receptive field, SAEM conducts self-attention enhancement operations on the image features, thus the global spatial information can be obtained through simple query and assignment.

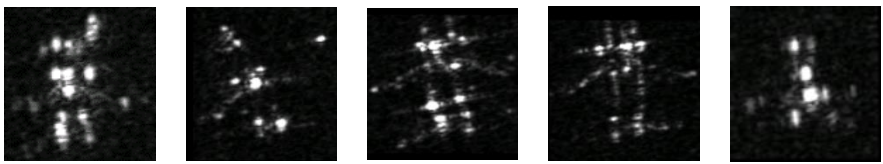

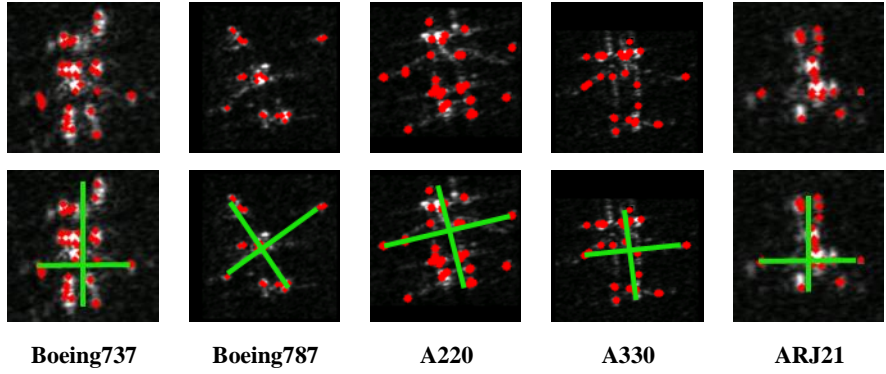

      **Boeing737**      **Boeing787**      **A220**      **A330**      **ARJ21**

**Fig. 3.** Extraction results of SSPs and size information. The first row shows the initial images, the second row shows the extraction results of SSPs, the third row shows the extracted size information.

### 1) Strong Scattering Point (SSP) extraction

Because of the special aircraft structure and complicated SAR imaging mechanism, radar wave is likely to form strong backscattering in the aircraft with dihedral angle, trihedral angle and other strong scattering structure (such as nose, tail, engine, wing and fuselage junction, etc.). The backscattering will appear as bright areas in SAR images and SSPs will not change greatly with external conditions. Therefore, SAR aircraft appears as discrete scattering points with weak integrity. Harris-Laplace detector has good detection invariance for rotation angle change, its algorithm has remarkable robustness[34]. Thus, in this article, Harris-Laplace detector is employed to extract SSPs in the aircraft slices. The results of SSPs extraction are shown in the second row of Fig. 3. On account of the aircraft's characteristic of axial symmetry, the least square method is adopted to carry out linear regression for the SSPs to extract the straight line of the fuselage. The least square method's particular computation is as follows: Assume that the extracted SSPs are expressed as $[(x_i,y_i), i \in \{1,2,\dots,n\}]$, the center of the sample points $(\bar{x},\bar{y})$ is calculated by

$$\bar{x} = \frac{1}{n}\sum_{i=1}^{n}x_i, \quad \bar{y} = \frac{1}{n}\sum_{i=1}^{n}y_i \quad (1)$$

Then calculate the intercept and the slope of the line to get the line through the fuselage

$$b = \frac{\sum_{i=1}^{n}x_i y_i - n\bar{x}\bar{y}}{\sum_{i=1}^{n}x_i^2 - n\bar{x}^2} \quad (2)$$

$$k = \bar{y} - b\bar{x} \quad (3)$$

The fuselage length can be expressed by the length between the two points of maximum distance on the line. Then extract the longest line segment in the vertical direction of the fuselage. The lengths of two segments represent the fuselage length and wingspan width of the aircraft respectively, as shown in the third row of Fig. 3.

### 2) Size Information Extraction Branch (SIEB)

As shown in Fig. 2, the proposed TSFE-N adds a size information extraction branch (SIEB) on the basis of the original image path from convolutional classification network. The orange part (CNN_SAEM) in Fig. 2 is constructed by convolutional neural network (CNN) for image feature extraction, and the lower branch shows the process of multi-modal feature extraction by multilayer perceptron (MLP). Then the target's size information features of text mode is updated to image features by adaptive projection.

SIEB is constructed mainly through MLP. SIEB takes the aircraft's fuselage length and wingspan width obtained from SSPs extraction mentioned above as input. The length and the width size information are standardized into the interval of [-1,1], and are connected according to the channel

$$\hat{\mathbf{x}}_e = \text{Concat}(\{length, width\}) \quad (4)$$

Where, length and width denote the target's length and width respectively, Concat($\cdot$) represents the concatenation of channels, $\hat{\mathbf{x}}_e \in \mathbb{R}^2$ denotes the size information intermediate encoding results. And then map size information $\hat{\mathbf{x}}_e$ to $\mathbb{R}^4$ through

$$\mathbf{x}_e = [\sin(\pi\hat{\mathbf{x}}_e), \cos(\pi\hat{\mathbf{x}}_e)] \quad (5)$$

Here sin($\cdot$) and cos($\cdot$) represent the sine and cosine function. After that, MLP network is used to obtain the domain knowledge features of text pattern, as

$$\mathbf{z}_e = \text{ReLU}(\text{LN}(f(\mathbf{x}_e))) \quad (6)$$

Where, ReLU($\cdot$) stands for ReLU activation function, LN($\cdot$) stands for layer normalization, and $f(\cdot)$ represents fully connected layer.

The multi-modal feature extraction branch dynamically updates the size information features represented by text into the image features by means of adaptive projection. In this way, the target features with domain knowledge weights are obtained: The weight of dynamic projection W is generated according to the characteristics of multi-modal domain knowledge $\mathbf{z}_e$, as

$$\mathbf{W} = \text{Reshape}(f(\mathbf{z}_e)) \quad (7)$$

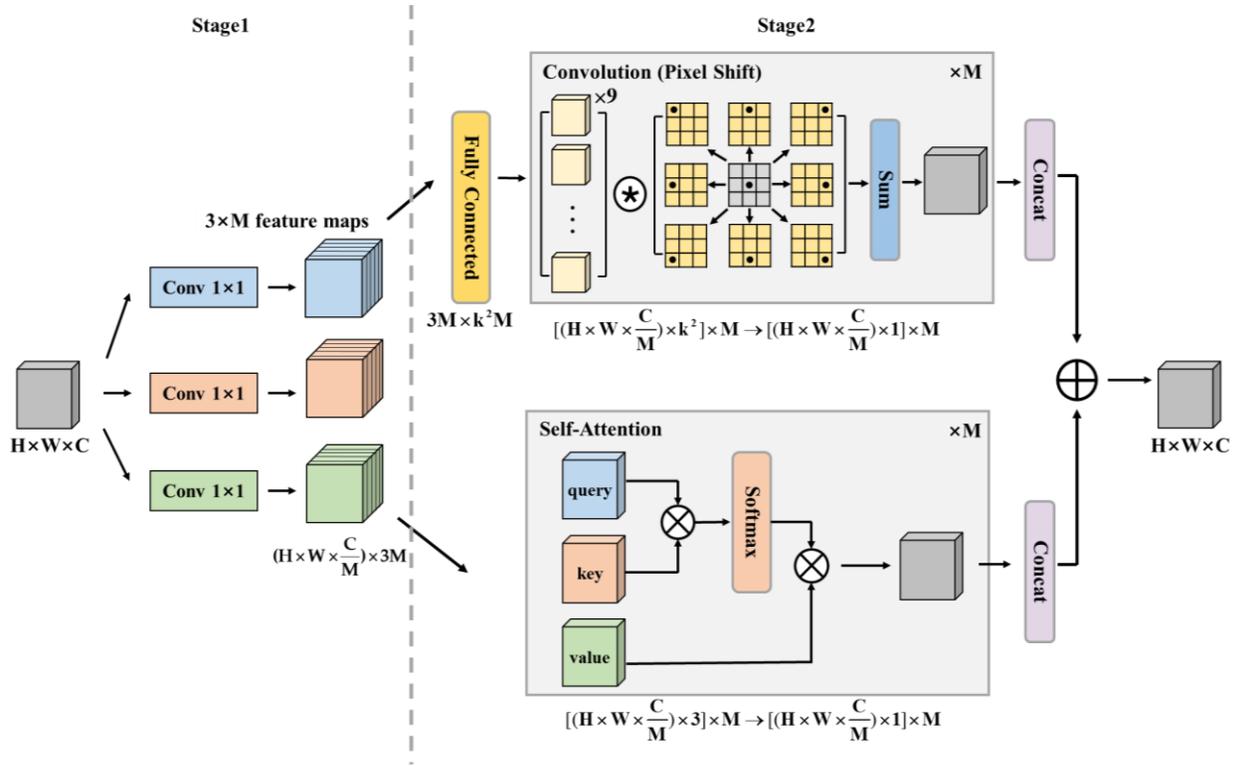

**Fig. 5.** Structure of the SAEM.

Where, Reshape(·) denotes that one-dimensional features are reconstructed into two-dimensional features, $f(·)$ represents the fully connected layer. Finally, by the means of adaptive projection, the multi-modal domain knowledge features are updated into the image features to obtain the target features

$$\mathbf{z}_i^{n+1} = \text{ReLU}(\text{LN}(f(\mathbf{z}_i^n; \mathbf{W}))) \quad (8)$$

Here, $\mathbf{z}_i$ represents the initial image features after global average pooling. $\mathbf{z}_i^n$ represents an image representation that is dynamically updated $n$ times, wherein $n \in \{1,2,…,N\}$. The above procedure uses $\mathbf{z}_i^0$ and $\mathbf{z}_e$ as the input and is iterated $n$ times until it completes the $N$ recursion.

### 3) Self-Attention Enhancement Module (SAEM)

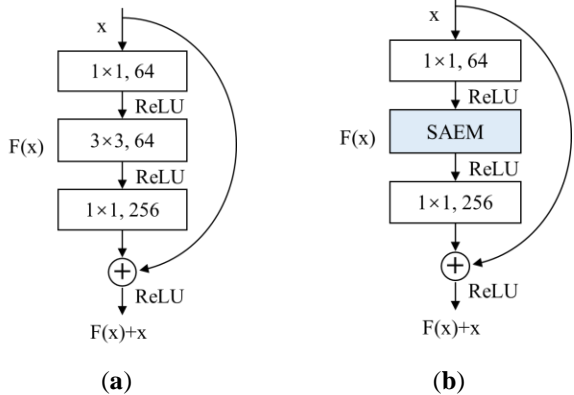

**Fig. 4.** Application of SAEM in the network. (a) Bottlenet structure before improvement; (b) Bottlenet structure after the addition of SAEM.

The image features are enhanced by self-attention module. Self-attention is a kind of dynamic and spatially variable operation, which is a kind of high-order global operation. It aims to improve model performance by building long-distance semantic interactions. The self-attention enhancement module (SAEM) is created in this section. The following takes ResNet-50 as an example to illustrate the network architecture of image feature extraction with SAEM added. The only difference from the original ResNet-50 is the replacement of the 3×3 convolutions in the ResNet-50's Bottlenet structure with our SAEM. The Bottlenet structures before and after the improvement are shown in Fig. 4.

As shown in Fig. 5, SAEM has two periods. In period one, the input features are expanded 3 times by three 1×1 convolution projections, the three generated feature maps are reshaped into M fragments respectively, and 3×M intermediate feature maps are obtained. In period two, SAEM integrates two paradigms, features are used through convolution and self-attention, in which a shared feature transformation structure is adopted:

#### a) Convolution Part

Firstly, the number of channels is expanded through the fully connected layer to generate $k^2$ feature maps, which are corresponding to multiple points in different offset directions. This can be understood as separating the convolution kernel of $k \times k$ in traditional convolution into $k \times k$ convolution kernels of 1×1. Then, spatial migration and aggregation operation are performed to obtain the results of convolution branches. This way of collecting information from local receptive fields is similar to the case with traditional convolution methods. Spatial migration is the shift and dot product of elements. In Fig. 5, $k=3$ is taken as an example to associate the shift operation with the convolution kernel, which can improve the capacity of the model on the basis of maintaining the ability of the original shift operation. The standard convolution computation can be represented as:

$$g_{ij} = \sum_{m,n} K_{m,n} f_{i+m-\lfloor k/2 \rfloor, j+n-\lfloor k/2 \rfloor} \quad (9)$$

where, $K_{m,n}$, $m,n \in \{0,1,\ldots,k-1\}$ represents the weight at position $(m,n)$ relative to the center of a $k \times k$ convolutional kernel. $f_{ij}$ is the feature at position $(i,j)$ in the input feature map, and $g_{ij}$ is the feature at position $(i,j)$ in the output feature map after convolution. For convenience, we assume that

$$g_{ij}^{(m,n)} = K_{m,n} f_{i+m-\lfloor k/2 \rfloor, j+n-\lfloor k/2 \rfloor} \quad (10)$$

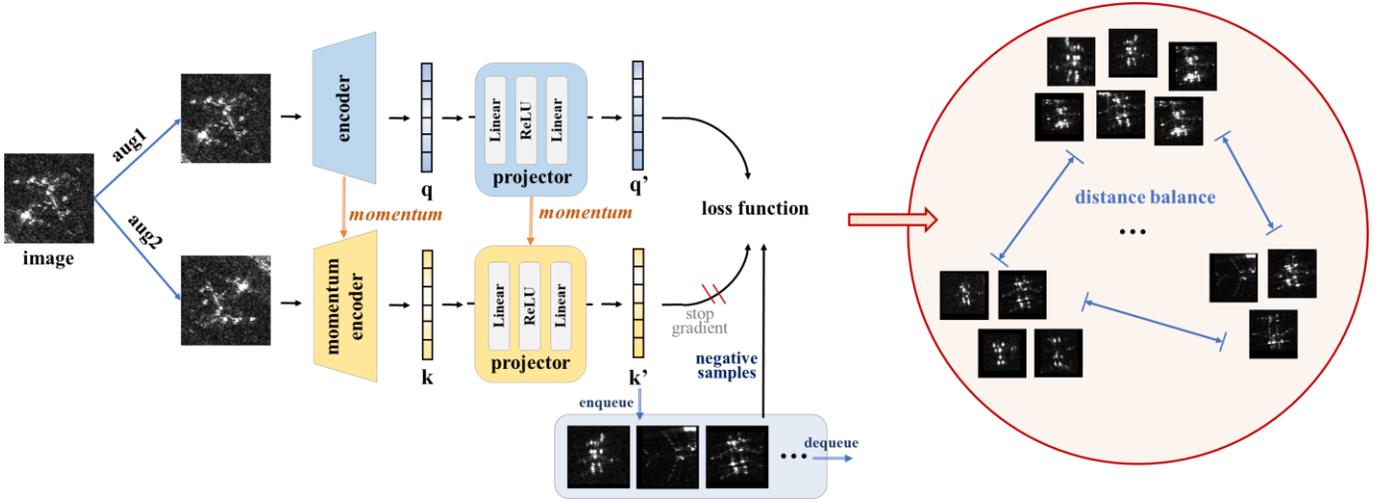

**Fig. 6.** Structure of MoCo v2.

Therefore, (9) can be rewritten as

$$g_{ij} = \sum_{m,n} g_{ij}^{(m,n)} \quad (11)$$

To further simplify the expression, we define

$$f_{i+\Delta x, j+\Delta y} \triangleq \text{Shift}(f_{ij}, \Delta x, \Delta y) \quad (12)$$

Therefore, spatial migration and aggregation operation can be expressed as

$$\begin{aligned} g_{ij}^{(m,n)} &= K_{m,n} f_{i+m-\lfloor k/2 \rfloor, j+n-\lfloor k/2 \rfloor} \\ &= \text{Shift}(K_{m,n} f_{ij}, m-\lfloor k/2 \rfloor, n-\lfloor k/2 \rfloor) \end{aligned} \quad (13)$$

$$g_{ij} = \sum_{m,n} g_{ij}^{(m,n)} \quad (14)$$

b) Self-attention Part

Inspired by Transformer [44], for the self-attention mechanism of images, images are viewed as a $w \times h$ sequence and intermediate features are collected into $M$ groups, where each group is an input vector. Each group contains three partial features, where each $1 \times 1$ convolution corresponds to one of them. Three feature maps are used as queries, keys and values, following the traditional self-attention module. The calculation is expressed as follows

$$q_{ij}^{(l)} = W_q^{(l)} f_{ij}, \quad k_{ij}^{(l)} = W_k^{(l)} f_{ij}, \quad v_{ij}^{(l)} = W_v^{(l)} f_{ij} \quad (15)$$

$$g_{ij} = \|_{l=1}^{M} \left( \sum_{a,b \in \mathcal{N}_k(i,j)} \left( \text{softmax}_{\mathcal{N}_k(i,j)} \left( \frac{(q_{ij}^{(l)})^T (k_{ab}^{(l)})}{\sqrt{\dim}} \right) \right) v_{ab}^{(l)} \right) \quad (16)$$

Where $f_{ij}$ is the feature at input feature map $(i,j)$. $W_q^{(l)}$, $W_k^{(l)}$, $W_v^{(l)}$ are the projection matrixes of query, key and value. $\mathcal{N}_k(i,j)$ represents a local pixel area with space extent $k$ as the center of $(i,j)$. dim is the dimension of $q_{ij}^{(l)}$, by calculating the division of dimension, the range of dot product can be normalized, thus controlling the variance of variables and ensuring the stability of gradients during the training process. $\|$ is the concatenation of M output vectors. Finally, the two paths are integrated and added

$$F_{out} = \alpha F_{att} + \beta F_{conv} \quad (17)$$

The coefficients $\alpha$ and $\beta$ are learnable parameters.

*B. Contrastive Self-Supervised Learning (CSSL)*

The problems of high cost of annotated data and uneven distribution of dataset are solved by a CSSL method. In this section, the CSSL structure for aircraft classification in SAR images is introduced, its basic principle is studied, and which contrast learning framework can effectively improve the accuracy of aircraft classification network is discussed.

SSL means to construct its own supervisory information on large-scale unsupervised data by a pretext task, and then train the network through the constructed supervisory information so that it can learn valuable feature representations for downstream tasks. It can be regarded as an unsupervised learning method with a special form of supervision. According to the experimental results in SimSiam [42], MoCo v2 [45] has the best transfer effect in downstream tasks, so the CSSL framework based on MoCo v2 is used in this study, as shown in Fig. 6. For one image in the current batch, the query and its corresponding key are coded to make up positive sample pairs. The negative samples are originated from the queue. MoCo v2 framework contains following steps.

1) **Pre-processing**

Data enhancement is a common way to implement pretext tasks. In a set of data, samples obtained from two different enhancements of the same input make up positive pairs, while those obtained from different inputs form negative pairs. Unlabeled data X and Y are given to input, the sample obtained by random transformation of unlabeled data X is the positive sample X+ of X, and the sample obtained by random transformation of unlabeled data Y is the negative sample X1- of X. Strong data enhancement strategy is used in the experiment, which is gaussian deblur strategy.

### 2) Feature extraction encoder

The transformed samples are inputted into the feature extraction encoder $f(\cdot)$ for feature extraction, and the corresponding feature are obtained. The encoder of MoCo v1 is ResNet-50. The encoder of MoCo v2 is ResNet plus MLP, which extends the linear fully connected layer after the convolutional layer in the MoCo v1 network structure into two nonlinear layers of MLP, and activates the function by ReLU. In this article, the proposed TSFE-N is used as the encoder. Momentum encoder k and encoder q have the same network structure. For parameter updating, the parameter of $f_q$ is expressed as $\theta_q$ and the parameter of $f_k$ is expressed as $\theta_k$. Only the parameter $\theta_q$ is updated by back propagation. $\theta_k$ is updated by momentum encoder

$$\theta_k \leftarrow m\theta_k + (1-m)\theta_q \quad (18)$$

Momentum coefficient $m \in [0,1)$. If a large momentum $m$ is selected, that is, m approaches 1, the momentum encoder updates very slowly and is less dependent on the current input. Momentum updates in (18) make the evolution of $\theta_k$ more stable than $\theta_q$. Thus, although the keys in the queue from different small batches are encoded by different encoders, the differences between these encoders can be small.

### 3) Loss function

Neural networks for fine-grained classification tasks usually use cross-entropy loss during training, because the differences between classes are more pronounced than those within a class. Cross-entropy loss enables deep networks to learn discriminative general features from large amounts of data, and minimize training errors. However, neural networks tend to focus on the commonalities of intra-class features, which may lead to overfitting, resulting in a significant drop in accuracy on the testing set compared to the training set. In order to address the overfitting problem while maintaining accurate classification, a similarity perception loss (SP loss) is designed to balance different categories and improve the network's generalization ability.

On the one hand, after the corresponding features are obtained by the encoder, the distance among positive samples is narrowed continuously and the distance among negative samples is enlarged successively by comparing loss function. The learning rate of cosine attenuation is adopted, and the loss function used in the calculation of loss is InfoNCE [46]

$$\mathcal{L}_{infoNCE} = -\log \frac{\exp(q \cdot k_+ / \tau)}{\sum_{i=0}^{K} \exp(q \cdot k_i / \tau)} \quad (19)$$

This loss function is the logarithmic loss of a $K+1$ softmax classifier, with $K$ negative samples and 1 positive sample. InfoNCE calculates the similarity of $q$ and $k$ by dot product. In (19), $k_+$ refers to the positive vector encoded by momentum encoder, $k_i$ refers to all positive and negative sample features. There is only one positive sample in control samples, the rest are negative samples. The denominator $\tau$ is the temperature parameter of softmax, which controls the sharpness and smoothness of the probability distribution. It can be seen from (19) that only the distance among positive sample pairs is calculated in the numerator, and the distance sum among positive and negative sample pairs is calculated in the denominator. When the positive sample pair distance is smaller, the negative sample pair distance is larger, and the loss is smaller.

On the other hand, considering that the conditional probability distribution of $N$-class targets is an element on $\mathbb{R}^N$, the Euclidean distance (L2 distance) is introduced to regulate overfitting and bring the distance between two sets of feature points closer. For the inputs $x_1$ and $x_2$ of the network, Euclidean metrics can be represented as

$$\begin{aligned} &\mathbb{D}_{EC}\left(p_\theta(y|x_1), p_\theta(y|x_2)\right) \\ &= \sum_{i=1}^{N} \left(p_\theta(y_i|x_1) - p_\theta(y_i|x_2)\right)^2 \\ &= \|p_\theta(y|x_1) - p_\theta(y|x_2)\|_2^2 \end{aligned} \quad (20)$$

In $N$-class target classification tasks, assuming the number of objects in the $i$-th class is denoted by $m_i$, the Euclidean metric between class $i$ and class $j$ can be formulated as the mean of Euclidean distances between all pairs of points across the two categories. Specifically, let $S_i$ denote the set of conditional probability distributions of all training points in the $i$-th class within the model (whose parameters are denoted as $\theta$)

$$S_i = \{p_\theta(y|x_1^i), p_\theta(y|x_2^i), \ldots, p_\theta(y|x_{m_i}^i)\} \quad (21)$$

Then, for this model, the Euclidean metric is

$$\begin{aligned} \mathbb{D}_{EC}\left(S_i, S_j; \theta\right) &\triangleq \frac{1}{m_i m_j} \left( \sum_{u,v}^{m_i, m_j} \mathbb{D}_{EC}\left(p_\theta(y|x_u^i), p_\theta(y|x_v^j)\right) \right) \\ &= \frac{1}{m_i m_j} \left( \sum_{u,v}^{m_i, m_j} \|p_\theta(y|x_u^i) - p_\theta(y|x_v^j)\|_2^2 \right) \end{aligned} \quad (22)$$

Finally, the similarity perception loss (SP loss) designed in this article is the sum of two parts of losses:

$$\mathcal{L} = \mathcal{L}_{infoNCE} + \mathbb{D}_{EC} \quad (23)$$

SP loss can optimize feature representations and improve the discriminative and generalization ability of the network by considering the similarity distance between positive and negative sample pairs. During the training process, cross-entropy loss and Euclidean metric loss are both calculated between the two branches of contrastive learning. When the perception loss reaches convergence, inter-class distance balance is achieved, leading to accurate classification and strong network generalization ability.

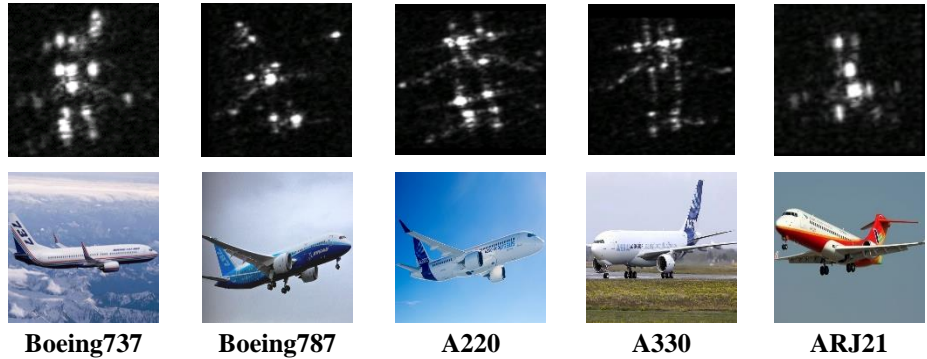

**Fig. 7.** Some types of SAR aircraft slices and their optical images in the Gaofen-3 dataset.

## IV. EXPERIMENTS AND ANALYSIS

A large quantity of experiments using data from the Gaofen-3 satellite and the TerraSAR-X satellite are carried out in this section to verify the validity of the proposed MS-Net. Firstly, the dataset is introduced. Secondly, the evaluation metrics of aircraft classification are introduced. Thirdly, the detailed parameters of the experiment are described. Finally, the experiment of this article is divided into two parts, and the performances of different components are analyzed in detail. The first part is to verify the validity of TSFE-N. The proposed model is compared with several advanced fine-grained classification methods, and the ablative performances of SIEB and SAEM are verified. The second part is the evaluation of the CSSL method, using a variety of self-supervised feature extractors for SAR image aircraft classification. Then, the proposed TSFE-N is transferred to the task of self-supervised aircraft classification network in SAR images, replacing the original pre-training feature extractor.

### A. Dataset Description

To address the issue of the insufficient open-source aircraft fine-grained classification datasets, SAR aircraft dataset from Gaofen-3 satellite and TerraSAR-X satellite is established. The Gaofen-3 satellite captures SAR images with C-band HH polarization in spotlight mode [34], [47]. The image slice sizes include 600×600, 1024×1024 and 2048×2048, with a total of 2000 images and a resolution of 1m. The TerraSAR-X satellite images contain a total of 53 large-scene SAR images, involve 10 airfields with a resolution of 1m. The mixed dataset contains 17 types of aircraft in six categories, including passenger aircraft, fighter aircraft, transport aircraft, refueling aircraft and so on. Fig. 7 shows a part of civil SAR aircraft slices and their optical images in this Gaofen-3 dataset. Fig. 8 shows the size distribution of various aircrafts. Fig. 9 shows the number of aircraft slices by type. The data analysis indicates that the size of different types of aircraft varies widely, and the aspect ratios of most of the targets are between 0.75 and 1.75. In order to retain the original size information of the aircrafts and facilitate subsequent processing, the original label boxes are considered as the benchmark to extend 5 pixels outward, the target is cut from the original image and zero edge filling is conducted. The sizes of all aircraft slices are adjusted to 320×320 pixels. A total of 6007 slices of aircrafts with a fuselage length and width of about 10-75m are obtained in the dataset. Since the SAR image of Gaofen-3 and TerraSAR-X satellites have different working frequency bands, polarization modes, interference performance and other characteristics, so there may be differences in image intensity. To solve this problem, two data sets are normalized to eliminate differences in image intensity, which was normalized to the range [0,255]. This method is helpful to maintain the consistency and comparability of data, so as to improve the accuracy of classification model.

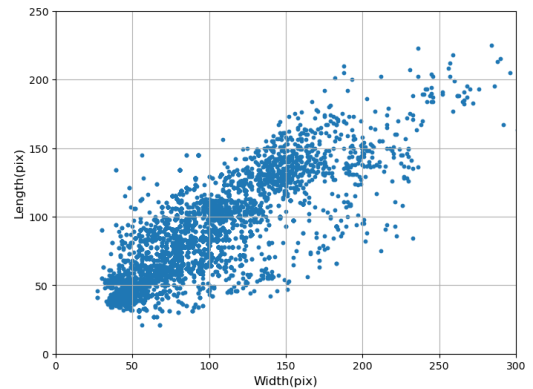

**Fig. 8.** Distribution of aircrafts' length and width.

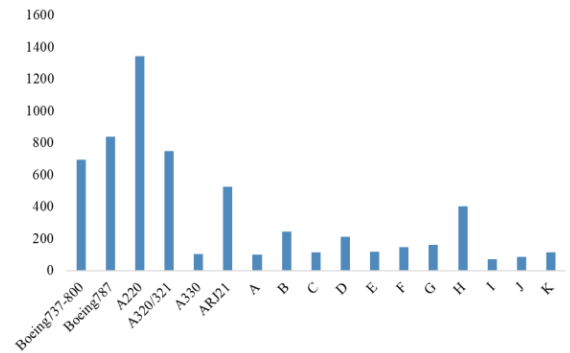

**Fig. 9.** The number of slices of each type of aircraft obtained from large-scene images.

### B. Evaluation Metrics

In the classification experiment, accuracy and confusion matrix are used to quantitatively evaluate the classification performance, heat map and T-SNE [48] method are used to qualitatively evaluate the classification performance. Accuracy is the most commonly used performance evaluation index in target classification algorithm. It represents the proportion of the number of samples correctly classified by the model to the total number of samples. Confusion Matrix (CM) is a visualization tool that reflects the classification performance of models, which is especially suitable for supervised learning. By comparing the measured position and category of the sample with the position and category of the true value of the sample, the matrix is determined, and the classification performance of each category is more refined.

TABLE II
THE NUMBER OF DATA USED IN THE TRANSFER EXPERIMENT.

| | | Boeing 737-800 | Boeing 787 | A220 | A320 /321 | A330 | ARJ 21 | A | B | C | D | E | F | G | H | I | J | K | total |
|---|---|---|---|---|---|---|---|---|---|---|---|---|---|---|---|---|---|---|---|
| Upstream unsupervised pre-train | | 484 | 587 | 940 | 522 | 72 | 366 | 68 | 169 | 80 | 147 | 81 | 101 | 112 | 281 | 48 | 60 | 78 | 4196 |
| Downstream linear protocol | 10% | 48 | 58 | 94 | 52 | 7 | 36 | 6 | 16 | 8 | 14 | 8 | 10 | 11 | 28 | 4 | 6 | 7 | 413 |
| | 20% | 96 | 117 | 188 | 104 | 14 | 73 | 13 | 33 | 16 | 29 | 16 | 20 | 22 | 56 | 9 | 12 | 15 | 833 |
| | 50% | 242 | 293 | 470 | 261 | 36 | 183 | 34 | 84 | 40 | 73 | 40 | 50 | 56 | 140 | 24 | 30 | 39 | 2095 |
| Test | | 208 | 252 | 404 | 225 | 32 | 158 | 30 | 73 | 35 | 64 | 36 | 44 | 48 | 121 | 21 | 26 | 34 | 1811 |

### C. Experimental Settings

All experiments are conducted in Ubuntu 20.04 system and TITAN RTX (24GB) GPU under Pytorch1.8.1 deep learning framework. The size of the training image is 320×320 pixels. Before training, random rotation, random crop, flipping and other standard practice for data enhancements are carried out on the samples, which are available in the torchvision package of PyTorch to improve the model's ability to generalize. For validation, a 224×224 pixels center crop is applied for evaluating the accuracy.

In experiments on TSFE-N, the model training process utilizes Stochastic Gradient Descent (SGD) [49] as the optimizer, with the momentum factor of 0.9. The initial learning rate is 0.01. To maintain a better gradient, the learning rates in the first 5 epochs are warmed up. Each experiment is conducted for 400 epochs, and the best evaluated models on the training set are saved. In the experiment, the dataset is divided into training set and test set in a ratio of 7:3, and the data of the training set and test set are equally distributed.

In experiments on CSSL, the same model training optimizer is SGD, and the momentum factor size is 0.9. The self-supervised network upstream feature extraction network training epoch=300, momentum coefficient $m$=0.999, the temperature parameter $\tau$=0.07. In the downstream linear classification task, the conventional linear protocol [38] is used for testing, which means that the parameters of the backbone (used for feature extraction) are frozen, and only the parameters of the fully connected layer (classification head) used for classification are fine-tuned. This method can demonstrate the quality of feature learning in the main network. The epoch in the linear protocol testing phase is set to 200. In addition, to address the problem of insufficient annotated sample data, relevant experiments are conducted in the downstream task of reducing the number of labels. The upstream training task on the self-supervised network uses 4,196 unlabeled training set data. In the downstream linear classification task, 10%, 20%, 50% and 100% of the training set data are respectively selected for training, and the best performing model in evaluation is retained as the final training result model. The test set consists of all 1,811 data, and the specific quantity information of the dataset is shown in Table II.

### D. Experimental Results and Analysis

**1) Experiments on TSFE-N**

*a) Comparison with State-of-the-Art Algorithms*

In order to assess the effectiveness of TSFE-N for fine-grained classification of SAR aircraft, the accuracy comparison results of the proposed model with several state-of-the-art image classification methods are displayed in Table III. For the sake of fair evaluation, the default settings of the original article are adopted, the single 224×224 crop is for evaluation. The performance of the proposed method is superior to the existing deep learning methods. The classification accuracy of this TSFE-N on the dataset achieves a highly competitive 94.478%, which is 1.822% higher than that of ResNet-50 [50].

TABLE III
EXPERIMENTAL RESULTS OF AIRCRAFT FINE-GRAINED CLASSIFICATION BY DIFFERENT CLASSIFICATION ALGORITHMS. (1-CROP 224×224).

| Method | Reference | params (M) | FLOPs (G) | Acc(%) |
|---|---|---|---|---|
| ResNet-50 [50] | CVPR 2016 | 23.54 | 4.13 | 92.656 |
| DenseNet-121 [51] | CVPR 2017 | 6.89 | 2.83 | 90.698 |
| inception_v3 [52] | CVPR 2016 | 21.82 | 2.85 | 89.526 |
| inception_v4 [53] | AAAI 2017 | 41.17 | 6.15 | 91.089 |
| inception_resnet_v2 [53] | AAAI 2017 | 54.33 | 6.50 | 92.114 |
| SKNet-34 [54] | CVPR 2019 | 21.76 | 3.67 | 93.579 |
| MobileNet-v3 [55] | ICCV 2019 | 4.17 | 0.22 | 93.359 |
| EffNet-B0 [56] | ICML 2019 | 3.99 | 0.39 | 91.316 |
| EffNet-B7 [56] | ICML 2019 | 63.52 | 5.17 | 93.154 |
| EffNetV2-M [57] | ICML 2021 | 50.81 | 6.19 | 94.179 |
| ConvNeXt-T [58] | CVPR 2022 | 27.81 | 4.45 | 93.822 |
| ConvNeXt-S [58] | CVPR 2022 | 49.42 | 8.68 | 94.375 |
| **Our Method** | -- | 26.39 | 6.98 | **94.478** |

b) Ablation Study

In order to illustrate the network composition of TSFE-N, a set of ablation experiments are conducted to reveal the effects of various improvements, including the addition or removal of SIEB and SAEM as described in Section III.A. All works are trained by using SGD algorithm based on ResNet-18, ResNet-50 and ResNet-101 trunks. Detailed experimental results are shown in Table IV. Taking ResNet-50 as an example, when SIEB is only added, the experimental accuracy is increased by 1.160%. When SAEM is only added, the test accuracy is improved by 1.325%. When two modules are added at the same time, the classification accuracy of the aircraft is even improved by 1.822%. In addition, under the ResNet-18 baseline, the introduction of TSFE-N improved the classification accuracy by 1.855%; under the ResNet-101 baseline, the classification accuracy of the model is improved by 0.607%. The model consistently shows better performance when two modules are added together. The ablation results fully verify the effectiveness of the module proposed by TSFE-N. With the guidance of multi-modal information, the feature encoder is enabled to learn more effective features.

The confusion matrix of 17 test types is shown in Fig. 10, indicating that the proposed method can effectively distinguish aircraft categories. In addition to the first six types of civil aviation passenger aircraft, type A B, type C D E, type G H I, type F K and type J respectively belong to the same main category of the aircraft, and they are similar in shape and scale. These similar aircrafts with high similarity in size, shape and structure are more likely to be misclassified. For example, type H and type I are difficult to be classified. Type I can easily be misidentified as type H which has more training samples. When SIEB is added, the classification accuracy of type I increases from 48% (baseline) to 67%. Similarly, for similar aircrafts type C and type D, some type D aircrafts are misidentified as type C. After SIEB is added, the classification accuracy of type D increases from 72% (baseline) to 91%. The classification accuracy of a total of 8 types of aircrafts has been improved by SIEB. SAEM improves the classification accuracy of type B by 10% without significantly damaging the classification effect of other categories. There are also 8 types of aircrafts' classification accuracy has been improved. Finally, TSFE-N improves the classification accuracy among 17 types of aircrafts by 1.822%, indicating the degree of progress of each module.

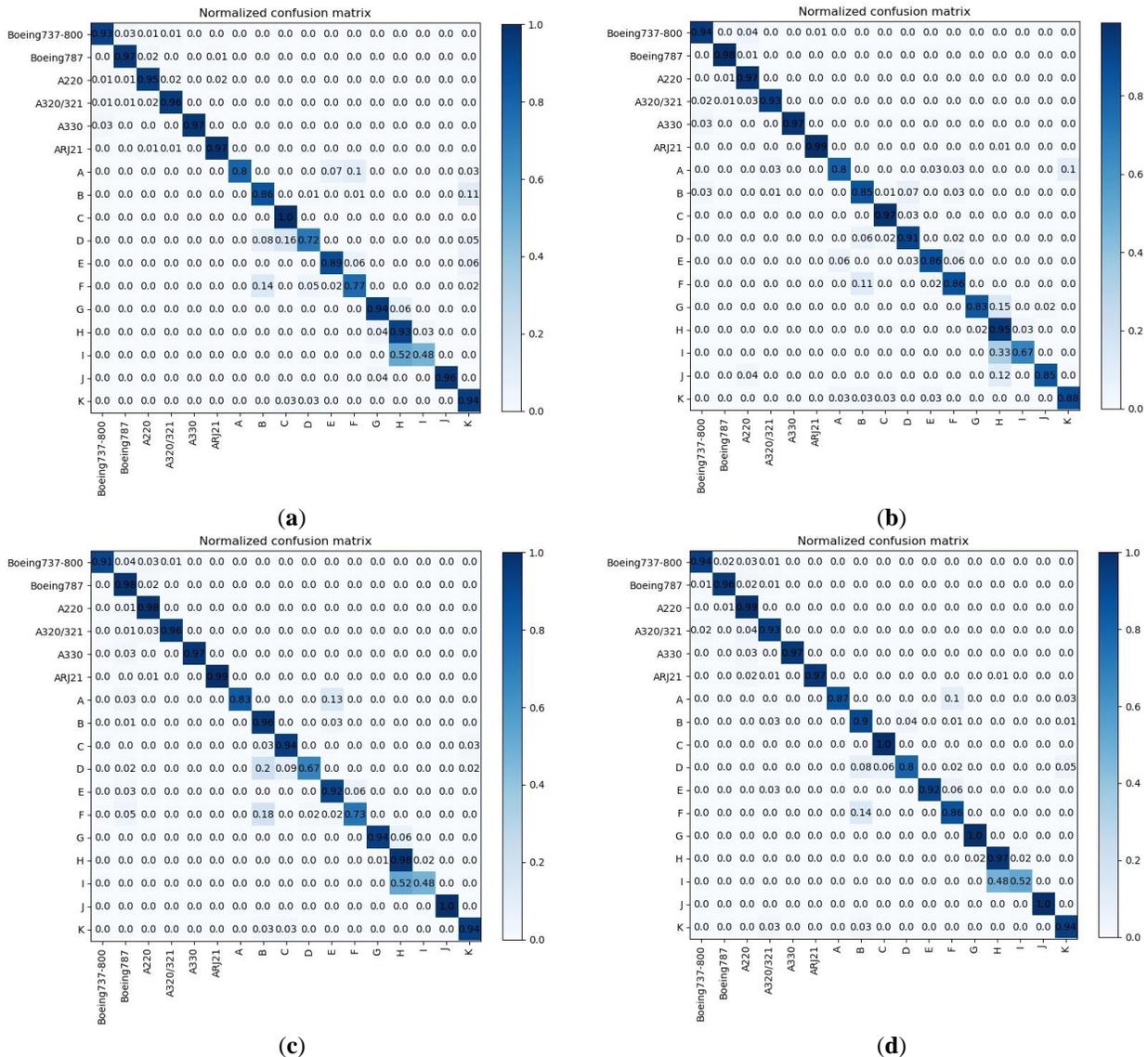

**Fig. 10.** Visualization results by confusion matrix. (**a**) Confusion matrix of baseline; (**b**) SIEB confusion matrix; (**c**) SAEM confusion matrix; (**d**) Confusion matrix of TSFE-N (both SIEB and SAEM).

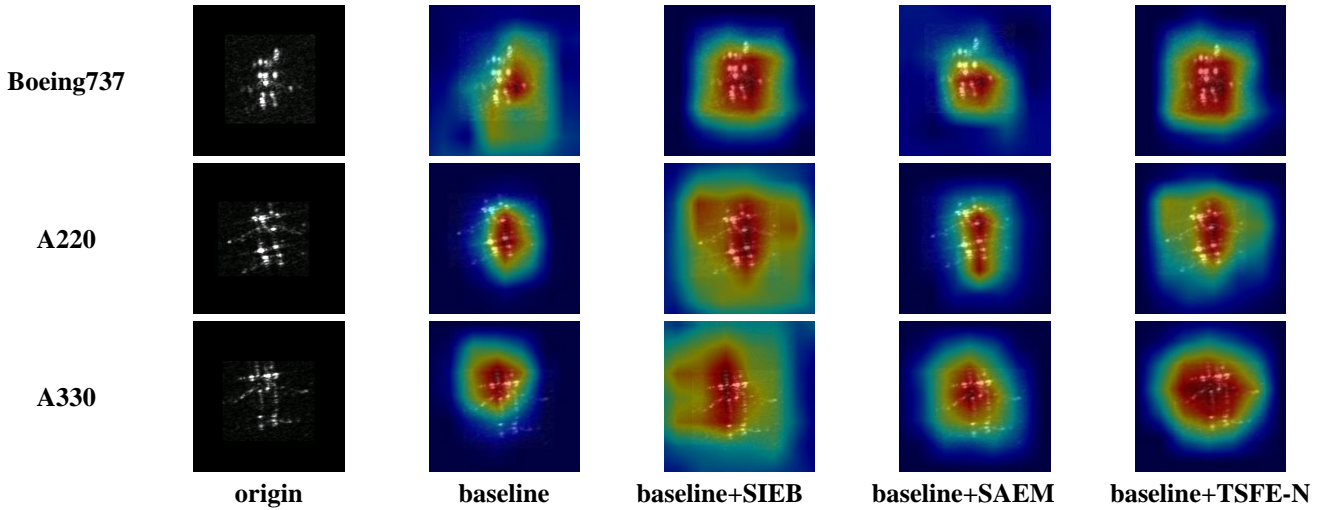

| | origin | baseline | baseline+SIEB | baseline+SAEM | baseline+TSFE-N |

**Fig. 11.** Heat maps of some experimental results. The first column shows the original images from Gaofen-3 satellite, the second column shows the results of baseline, the third and fourth columns are the results of SIEB and SAEM added respectively, the last column shows our results.

Table IV
ABLATIVE STUDY OF SIEB AND SAEM.

| Backbone | Method | | Acc(%) |
| --- | --- | --- | --- |
| | SIEB | SAEM | |
| ResNet-18 | × | × | 88.183 |
| | ✓ | × | 89.287 |
| | × | ✓ | 89.347 |
| | ✓ | ✓ | 90.038 |
| ResNet-50 | × | × | 92.656 |
| | ✓ | × | 93.816 |
| | × | ✓ | 93.981 |
| | ✓ | ✓ | 94.478 |
| ResNet-101 | × | × | 94.589 |
| | ✓ | × | 95.086 |
| | × | ✓ | 94.920 |
| | ✓ | ✓ | 95.196 |

To conduct a more comprehensive assessment of the model's efficacy, the original SAR aircraft slices and the extracted target length and width are taken as the input of the network, then the feature extracted after the last layer of convolution is visualized. The heat map visualization results of the algorithm are displayed in Fig. 11. It can be seen from the comparison that in the heat map of the proposed method, the coverage area of the key components of the aircraft is the most accurate, and the color of the edge of the target is darker. It indicates that SIEB can protect the scale information of the object, and SAEM can accurately enhance the significance of the core area of the aircraft while paying less attention to the background noise. It suggests that TSFE-N can make multi-mode features pay more attention to the global image and key areas during the transmission process, and bring higher accuracy to the fine-grained classification of aircrafts in SAR images.

The T-SNE [48] method is used to visualize image representations. Different types of aircrafts in the dataset are described in different colors. Fig. 12 indicates that the proposed method generates more distinctive image representations than the previous research. It proves that after the adoption of SIEB, the distinction between feature clusters of different categories is higher. Because SIEB converts image representation through weighted features generated by additional information, which involves interaction in a wider range of dimensions. Similarly, after the adoption of SAEM, the feature clusters of similar type are more compact, and the decision boundaries between different types are clearer. Because the self-attention module SAEM improves feature representation ability by constructing a long-distance semantic interaction, it is conducive to aircraft fine-grained classification. Therefore, the proposed TSFE-N method improves the distinguishing degree of features, which is more effective than the previous work.

c) Parameter Analysis

In (17), the SAEM module integrates two parameters $\alpha$ and $\beta$ to fuse two paths. $\alpha$ and $\beta$ reflecting the model's biases towards self-attention or convolution at different depths. Both $\alpha$ and $\beta$ are learnable parameters, they are bound to the model via the

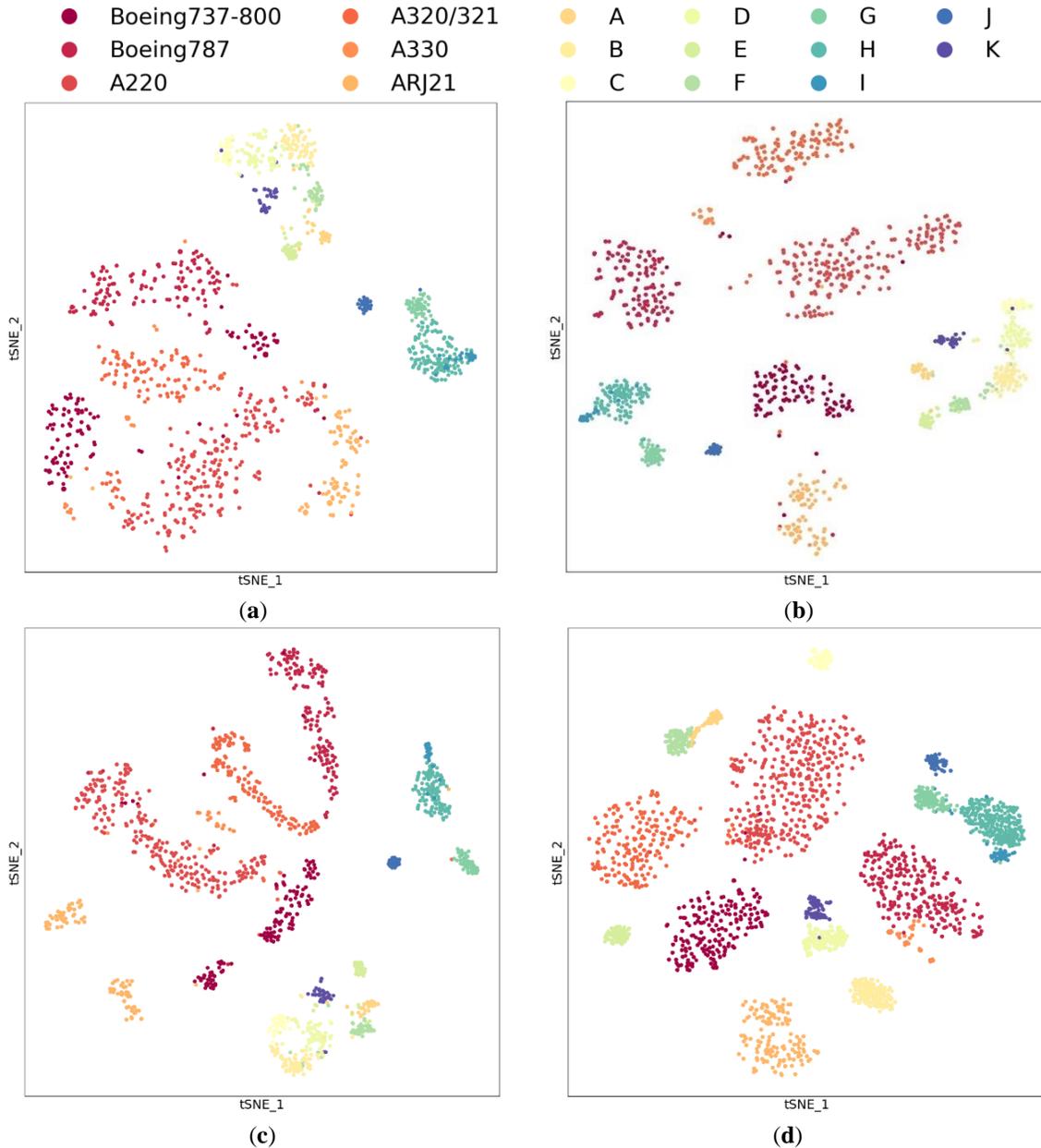

**Fig. 12.** Visualization of t-SNE representation under the well-trained model. **(a)** Visualization of Baseline; **(b)** SIEB visualization; **(c)** SAEM visualization; **(d)** Visualization of TSFE-N (both SIEB and SAEM).

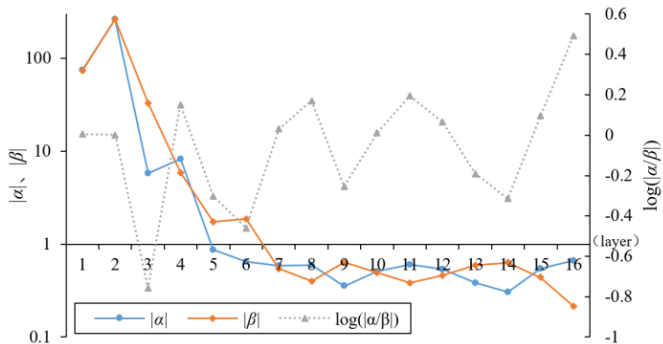

**Fig. 13.** The curves of the variations of $|\alpha|$, $|\beta|$ and $\log(|\alpha/\beta|)$ in different Bottleneck layers after the improvement of SAEM.

torch.nn.Parameter() function and are automatically updated with each iteration of the optimization process. We graphed the learned $\alpha$ and $\beta$ parameters from different Bottleneck layers after the improvement of SAEM as a line chart, respectively denoted as $|\alpha|$, $|\beta|$ and $\log(|\alpha/\beta|)$, as shown in Fig.13.

The curves of parameters $|\alpha|$ and $|\beta|$ respectively demonstrate the rate fluctuations of the self-attention and convolution paths, with less pronounced changes as the layers grow deeper. It suggests that deep models have more stable preferences as training progresses. Meanwhile, the curve of $\log(|\alpha/\beta|)$ clearly indicates the ratio between the two paths. When $\log(|\alpha/\beta|)<0$, convolution outperforms self-attention in feature extraction, while self-attention performs better than convolution otherwise. The experimental results indicate that the improved ResNet-50 model tends to use a combined learning of the two paths. Moreover, as the model becomes deeper, self-attention eventually outperforms convolution in terms of overall performance.

2) **Experiments on CSSL**

   a) *Comparison with State-of-the-art Self-supervised Algorithms*

The proposal and development of CSSL provides a solution to the problem that the supervised learning relies heavily on large-scale and high-quality labeled data. We evaluate the effectiveness of 5 types of CSSL networks including MoCo, MoCo v2, SimCLR, BYOL, and SimSiam for aircraft fine-grained classification in SAR images. The performances of several self-supervised methods are compared, params and FLOPs are used to describe the spatial complexity and time complexity of the model. The experimental results are listed in Table V. MoCo v2 is based on MoCo v1 with the adoption of MLP and cosine learning rate, and has stronger data enhancement (gaussian deblur). The linear classification results show that MoCo v2 achieves the best accuracy of 86.969% (300-epoch unsupervised pre-training, 200-epoch downstream training) among the methods using negative samples. It can be speculated that the SAR enhancement strategy used by MoCo v2 greatly improves the linear classification accuracy of aircraft in SAR images and enables the upper-reaches encoder to learn more effective features. Fig. 14 shows the training loss curve and the accuracy curve of the MoCo v2 in the downstream classification network. The loss gradually converges at the 70th epoch, and the overall convergence speed is fast. This indicates that the self-supervised contrast learning model gradually tends to be stable in the SAR aircraft classification training task and has good learning ability. The situations of the lack of labels will occur in the actual SAR aircraft classification task. It would be very practical to learn generic models from cognate images through CSSL and then migrate to downstream specific tasks. According to the results, the introduction of self-supervised pre-training network into SAR aircraft classification task can learn the effective features inside the image, and reduce the dependence of SAR data on expert annotation.

Table V
THE OVERALL ACCURACY OF DIFFERENT CSSL NETWORKS ON TEST SETS IS OBTAINED BY USING LINEAR CLASSIFIER (RESNET-50, 1-CROP 224×224).

| Method with negative samples or not | Method | params (M) | FLOPs (G) | Unsupervised pre-train batch | Unsupervised pre-train epoch | Acc (%) |
|---|---|---|---|---|---|---|
| Yes | SimCLR | 11.50 | 1.82 | 1024 | 400 | 82.917 |
|  | MoCo | 47.54 | 16.86 | 256 | 300 | 85.036 |
|  | **MoCo v2** | 55.93 | 8.27 | 256 | 300 | **86.969** |
| No | BYOL | 72.12 | 10.78 | 128 | 300 | 85.328 |
|  | SimSiam | 38.20 | 8.29 | 256 | 400 | 83.215 |

In addition, it is worth mentioning that for SimCLR network, since memory bank is not used, batch size is increased during training. The batch size is 1024 in the experiment. That is, for every positive sample, there will be 2046 negative samples. The SGD optimizer is unstable when training with large batch size, so LARS optimizer is used for SimCLR.

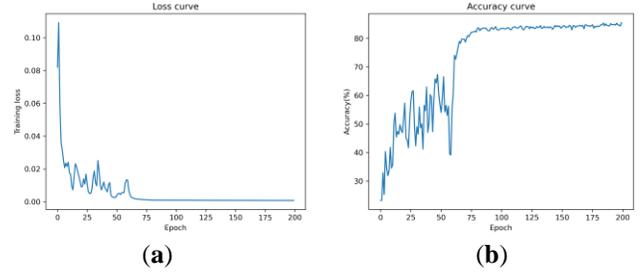

(a)  (b)

**Fig. 14.** The training loss curve and accuracy curve of linear classification.

In order to make a fair comparison, the results of unsupervised pre-training at different epochs are also compared (the training epoch of downstream linear classifiers is 200). The details are shown in Fig. 15. BYOL method without negative samples reached the highest accuracy of 88.268% (800-epoch unsupervised pre-training), outperforming MoCo v2's 87.897% (800-epoch unsupervised pre-training). Overall speaking, the MoCo network quickly achieves a good accuracy with a small number of params and a small computation.

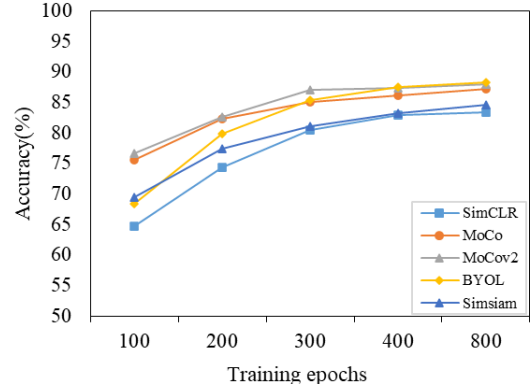

**Fig. 15.** The linear classifier accuracy of different self-supervised networks and different pre-training epochs.

   b) *Ablation Study*

In this section, the effectiveness of the proposed SPloss has been validated through several ablative experiments. As mentioned in section IV.C.a, all the work is based on the MoCo v2 network as the baseline network. The detailed experimental results are shown in Table VI. When the similarity perception loss (SPloss) was improved, the test accuracy increased by 0.993%. Comparing with the classification accuracy of other target classification algorithms in Table V, the proposed improved network performs better on the SAR aircraft classification.

Table VI
ABLATION STUDY OF SPLOSS.

| Method | Acc(%) |
|---|---|
| MoCo v2 | 86.969 |
| MoCo v2+SPloss | **87.962** |

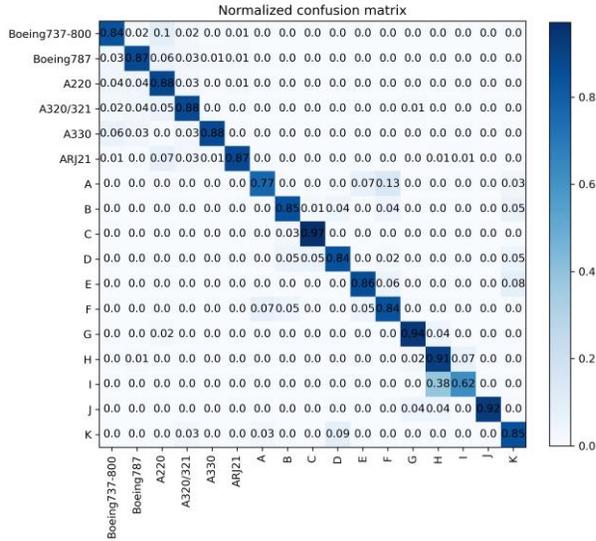
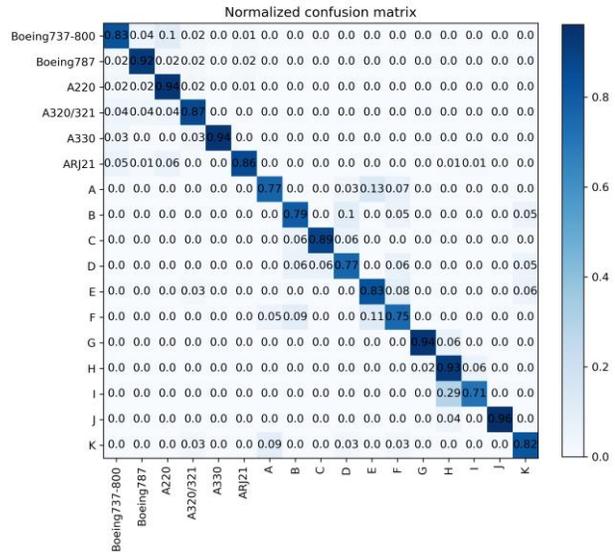

(a)           (b)

**Fig. 16.** Visualization results by confusion matrix. **(a)** Confusion matrix of baseline; **(b)** SPloss confusion matrix.

The confusion matrix of the 17 test categories is shown in Fig. 16. Similarly, among the aircraft types in the figure, the first six categories belong to civil airliners, type A B, type C, D, E, type G, H, I, and type F, K belong to the same aircraft category, they are very similar in appearance, size, and structure and are more likely to be misclassified by the classifier. By analyzing the confusion matrix, it can be seen that: (1) Self-supervised contrastive learning methods can alleviate the impact of long-tailed distribution in datasets to some extent. The results of the fully supervised network shown in Fig. 10(a) indicate that the categories with relatively few samples, such as classes D, F, and I, have poor classification performance, and the model has difficulty learning the feature representations of these classes. However, the results of the self-supervised network shown in Fig. 16(a) demonstrate that the classification accuracy of each type of object noticeably decreases due to the influence of the number of samples. Therefore, self-supervised contrastive learning methods can effectively improve the classification performance of the classifier for categories with relatively few samples in the dataset. (2) As can be seen from Fig.16(b), SPloss can provide a more effective measure of positive and negative sample distance for the model, and improve the training effect and network generalization ability of the contrast learning model by shortening the distance of the feature points set. This improvement improves the classification accuracy of type Boeing787, A220, A330, H, I and J. Compared with the baseline network, the overall test accuracy is increased by 0.993%.

   c) Performance of Self-supervised Algorithm in Downstream Transfer Tasks

Although ImageNet dataset in the optical scenario can provide some pre-training parameters, the performance may be affected by the significant differences between natural images and SAR remote sensing when transferred to downstream datasets. Furthermore, in practical SAR aircraft classification tasks, although there are increasing amounts of available SAR images, the cost of SAR image interpretation is expensive due to the need for expert knowledge, resulting in missing labels. As self-supervised pre-training does not require annotated data, this section explores the practical value of pre-training a universal model from homologous images through self-supervised learning and then transferring to specific downstream classification tasks.

Table VII
ABLATION STUDY OF SPLOSS.

| Method | The amount of fine tuning data with labels | Acc(%) |
|---|---|---|
| ResNet-50 | 10% | 56.322 |
|  | 20% | 57.261 |
|  | 50% | 71.452 |
|  | 100% | 92.656 |
| MoCo v2 | 10% | 62.507 |
|  | 20% | 67.366 |
|  | 50% | 83.932 |
|  | 100% (linear protocol) | 86.969 |
| our method (MoCo v2+SPloss) | 10% | 62.548 |
|  | 20% | 68.751 |
|  | 50% | 84.185 |
|  | 100% (linear protocol) | 87.962 |

In the training process of this section's experiment, all parameters of the upstream self-supervised network are pre-trained using the complete unlabeled training set. In the downstream classification task, only a portion of the labeled data is used to fine-tune the classifier parameters rather than all of the labeled data. The network's performance is still evaluated using the aforementioned "linear protocol" method, and the test dataset remains the same. The results are shown in Table VII.

As shown in the Table VII, the self-supervised contrastive learning pre-training network can effectively improve the accuracy of SAR aircraft classification as the fine-tuning data increases. When using the MoCo v2 self-supervised network as the baseline and fine-tuning with 50% and 100% of the downstream data, the accuracy only differs by 3.037%, while 2101 labeled data is saved. However, when training is performed in a fully supervised manner, the classification accuracy decreases by 21.204% when reducing the training data size from 100% to 50%, and decreases by 35.395% when further reducing the training data size to 20%. Therefore, it is evident that the self-supervised contrastive learning network can significantly reduce annotation costs while ensuring classification accuracy. Additionally, the self-supervised contrastive learning network can obtain feature representations through pre-training on unlabeled data and fine-tuning on different downstream tasks. It effectively addresses the problems of insufficient data and poor-quality labeling, thus improving network transfer learning efficiency and accuracy. This approach allows the network to achieve better performance and unlocks new opportunities for expanding application scenarios.

  d) *Performance of MS-Net*

The two-sided path feature extraction network (TSFE-N) is transferred to the SAR aircraft fine-grained classification task of self-supervised pre-training, replacing the original pre-training feature extractor. The classification results are shown in Table VIII. The results show that the proposed SIEB and SAEM increase the MoCo v2 self-supervised linear classification accuracy of the aircrafts by 0.784% and 0.533% respectively. The aircraft fine-grained classification accuracy of the targets in TSFE-N network reached 88.374%, which is 1.405% higher than that of the baseline network (MoCo v2). The overall aircraft fine-grained classification accuracy of the targets of our proposed MS-Net network reached 88.459%.

Table VIII
COMPARISON OF TSFE-N PERFORMANCE ON MOCO V2 NETWORK.

| Method | Acc(%) |
|---|---|
| MoCo v2 | 86.969 |
| MoCo v2+SIEB | 87.753 |
| MoCo v2+SAEM | 87.502 |
| MoCo v2+TSFE-N(SIEB+SAEM) | 88.374 |
| MoCo v2+SPloss | 87.962 |
| **our MS-Net**(MoCo v2+TSFE-N+SPloss) | **88.459** |

V. DISCUSSION

This section provides a deeper analysis of the proposed method and proposes potential research directions for future work.

In this article, a fine-grained aircraft classification architecture for SAR images named MS-Net is proposed.

**1)** TSFE-N is used to extract multi-modal knowledge features in SAR images, which includes two key components: SIEB and SAEM. SIEB updates the target's size information features of the text pattern to the image features through adaptive projection, which can protect the scale information of the objects. SAEM performs self-attention enhancement on image features, and improves model performance by constructing long-distance semantic interaction. SAEM can accurately enhance the significance and integrity of the aircrafts' core area. The design of SIEB and SAEM has improved the aircraft classification performance. However, while the proposed approach shows promising results, there remain some limitations that need to be addressed. The SIEB method is affected by the extraction accuracy of strong scattering points. Different sampling methods of different radars will lead to different sizes of the same aircraft, and the sensitivity of strong scattering points to the attitude of the target brought by different angles may lead to slight errors in the extraction of size information, which will affect the classification of the target model. In this article, the target size information is only used as auxiliary information to provide useful guidance for the fine-grained aircraft classification network based on the length, width and even the aspect ratio of the target, without requiring the network to extract the size information in fine detail. Later, physics knowledge can be integrated into the deep learning network to guide the classification network. And the target detection and recognition tasks can be combined to obtain more accurate target size information through the detection network. In addition to size information, more unique features of the aircraft (such as engine number) and geographic location information can also be introduced for precise network design.

**2)** CSSL network is adopted to reduce the dependence of the model on labeled samples and help to learn the initial feature information independent of labels from unbalanced data. A similarity perception loss function (SPloss) is designed to optimize the similar distance between positive and negative sample pairs, which helps to avoid overfitting phenomenon in the network and enhance the discriminant and generalization ability of the network. In the experiment, we evaluated the effectiveness of the classical self-supervised network in the SAR aircraft classification scenario, verified the rationality of the proposed SPloss, and proved that the self-supervised learning can effectively expand in the scenario with insufficient actual data through the transfer experiment. Future work will further consider the portability of multimodal knowledge between satellite data. In this direction, multi-source SAR data can be used for cross-domain research. The SAR data obtained under different observation conditions are often different. Combining multi-source SAR data can improve the accuracy and robustness of classification.

VI. CONCLUSIONS

A novel multi-modal self-supervised aircraft classification network (MS-Net) is proposed in this article. First, to fully exploit the latent impact of multi-modal information, a two-sided path feature extraction network (TSFE-N) is constructed. The size information of text mode is obtained by SSPs extraction. The size information extraction branch (SIEB) is designed to obtain the domain knowledge features of text mode, and protect the scale information of aircrafts. SAEM is created to refine the global information features of the aircrafts and enhance the saliency of the aircraft core area accurately. Then, the applicability of CSSL in remote sensing image classification is evaluated and a practical SPloss is presented. The experimental results indicate that the self-supervised network can effectively learn useful label-independent initial feature information from unbalanced data. The introduction of the self-supervised pre-training network into SAR aircraft classification can greatly reduce the labeling cost of remote sensing data and learn effective features inside the image, resulting in an improvement of the accuracy of downstream classification tasks. It is convinced that comparative learning has great promise in SAR aircraft classification. The approach indicates significant improvements over the baseline network on our own SAR aircraft dataset constructed from Gaofen-3 satellite and TerraSAR-X satellite images. It can be verified by a lot of experiments that MS-Net can effectively reduce the classification difficulty of aircrafts with similar appearance. In the case of no label, the proposed algorithm achieves a classification accuracy of 88.46% among 17 types of aircrafts, which has pioneering significance in the field of fine-grained classification of aircraft in SAR images.